%% file: main.tex
\documentclass{article} 
\usepackage[preprint]{neurips_2026}

\input{math_commands}

\usepackage[utf8]{inputenc} 
\usepackage[T1]{fontenc}    

\usepackage{url}            
\usepackage{booktabs}       
\usepackage{amsfonts}       
\usepackage{nicefrac}       
\usepackage{microtype}      
\usepackage{xcolor}         
\usepackage{hyperref}
\hypersetup{
	colorlinks=true,
	linkcolor=red,
	filecolor=blue,      
	urlcolor=magenta,
	citecolor=black,
}
\usepackage{stmaryrd}
\usepackage{xspace}
\newcommand{\method}{\textsc{TAPO}\xspace}

\usepackage{amsmath}
\usepackage{amssymb}
\usepackage{mathtools}
\usepackage{amsthm}
\usepackage{dsfont}
\usepackage{diagbox}
\usepackage{adjustbox}
\usepackage{multirow}
\usepackage{makecell}
\usepackage{wrapfig}
\usepackage{subfigure}
\usepackage{graphicx}

\usepackage{caption}
\usepackage{algorithm}
\usepackage{algpseudocode}

\usepackage{yhmath}
\usepackage{pifont}
\usepackage{tikz}
\usepackage{standalone}
\usetikzlibrary{shapes.geometric, arrows.meta, positioning, backgrounds, fit, calc, decorations.pathreplacing}

\theoremstyle{plain}

\theoremstyle{definition}

\theoremstyle{remark}

\usepackage[capitalize,noabbrev]{cleveref}
\usepackage[textsize=tiny]{todonotes}
\crefname{section}{Sec.}{Secs.}
\Crefname{section}{Section}{Sections}
\Crefname{table}{Table}{Tables}
\crefname{table}{Tab.}{Tabs.}

\title{Learning from Your Own Mistakes: Constructing Learnable Micro-Reflective Trajectories for Self-Distillation}

\author{
\textbf{Zhilin Huang}$^{1,2}$\quad \textbf{Hang Gao}$^{1}$\quad \textbf{Ziqiang Dong}$^{1}$\thanks{Corresponding Author}\quad \textbf{Yuan Chen}$^{1}$\quad \textbf{Yifeng Luo}$^{1}$\\
\textbf{Chujun Qin}$^{3}$ \quad \textbf{Jingyi Wang}$^{1}$ \quad \textbf{Yang Yang}$^{1}$ \quad \textbf{Guanjun Jiang}$^{1}$ \\[0.5em]
$^{1}$Qwen Business Unit of Alibaba $^{2}$Tsinghua University $^{3}$Peking University \\
\\
\texttt{\{henry.gh,ziqiang.dzq,duomo.cy,luoyifeng.lyf\}@alibaba-inc.com} \\
\texttt{\{huabuwan.wjy,chris.yang,guanj.jianggj\}@alibaba-inc.com} \\
\texttt{\{zerinhwang03,chujun\_qin\}@pku.edu.cn}\\
}

\begin{document}

\maketitle

\input{sections/00-abstract}
\input{sections/10-intro}
\input{sections/20-related}

\input{sections/30-preliminary}
\input{sections/40-methods}
\input{sections/50-experiments}
\input{sections/60-conclusion}
\input{sections/70-acknowledge}

\bibliography{ref}
\bibliographystyle{neurips_2024}

\end{document}

%% file: math_commands.tex
\usepackage{amsmath,amsfonts,bm}

\def\eqref#1{equation~\ref{#1}}

\def\1{\bm{1}}

\DeclareMathAlphabet{\mathsfit}{\encodingdefault}{\sfdefault}{m}{sl}
\SetMathAlphabet{\mathsfit}{bold}{\encodingdefault}{\sfdefault}{bx}{n}



%% file: sections/00-abstract.tex
\begin{abstract}

Self-distillation improves reasoning in large language models by using the model's own rollouts as training signal, typically through implicit logit-level alignment that minimizes KL divergence toward a privileged target distribution. However, because this supervision is generated via uncontrolled sampling, it provides no diagnostic insight into the model's specific errors or corrective guidance for its individual failure patterns. Consequently, the model learns to imitate a privileged distribution rather than receiving fine-grained corrections that pinpoint where and why its reasoning fails.
In this paper, we propose \textbf{T}rajectory-\textbf{A}ugmented \textbf{P}olicy \textbf{O}ptimization (\textbf{\method}), which advances self-distillation from implicit distributional alignment to explicit trajectory construction. During RL training, the model produces both correct and incorrect rollouts to the same query, and \method leverages this contrastive structure to construct \textbf{micro-reflective corrections}, new training trajectories that retain the model's erroneous reasoning up to the point of failure, then insert a natural-language diagnosis and corrected reasoning guided by a correct reference from the same sampling group. Since each trajectory is anchored in the learner's own prefix and solutions, the corrective signal preserves the model's on-policy distribution to a greater extent than the position-wise alignment imposed by KL-based methods. To integrate these trajectories, \method introduces difficulty-aware candidate selection at the model's capability boundary and decoupled advantage estimation to prevent gradient contamination. 
Experiments on AIME 2024, AIME 2025, and HMMT 2025 show that \method achieves consistent improvements over GRPO under the same number of training steps. 
Further analysis demonstrates that \method strengthens both first-pass reasoning and error-correction effectiveness.

\end{abstract}

%% file: sections/10-intro.tex
\section{Introduction}

Large Language Models (LLMs)~\citep{wang2026survey,chen2026towards,huang2023towards,xu2025toward,zhao2026survey,lv2024specfuse,patil2025advancing,zhou2025variational} have achieved remarkable progress on complex reasoning tasks through reinforcement learning with verifiable rewards (RLVR)~\citep{openai-o1,deepseek-r1,qwen3,yao2026rethinking,qi2026rethinking,wen2025reinforcement, mroueh2025reinforcement}. This paradigm, building on foundational work in chain-of-thought prompting~\citep{wei2022chain,yu2023towards,zhang2022automatic,mitra2024compositional} and process reward modeling~\citep{lightman2024lets,zhou2025reinforcing}, has enabled models to tackle increasingly challenging mathematical reasoning benchmarks~\citep{hendrycks2021math,cobbe2021gsm8k}. A central question in this paradigm is how to provide learning signal that effectively guides the model toward better reasoning. Outcome-level approaches such as Group Relative Policy Optimization (GRPO)~\citep{deepseekmath} offer sparse but stable signal by sampling multiple solutions and computing group-relative advantages. Self-distillation approaches such as On-Policy Distillation (OPD)~\citep{agarwal2024opd} and On-Policy Self-Distillation (OPSD)~\citep{zhao2026opsd,cui2026brief,lv2024taekd} take a richer approach that uses the model's own rollouts as the basis for generating dense token-level supervision, aligning the model's output distribution toward a privileged target distribution derived from correct solutions. Both paradigms have proven effective, yet they embody different philosophies of how a model should learn from its own reasoning.

Current self-distillation methods implement this principle through implicit distributional alignment. Whether the privileged target comes from an external teacher (OPD) or from the same model conditioned on a verified correct solution (OPSD), the model is optimized to minimize the KL divergence between its output distribution and the target at every token position. The supervision consists of continuous token-level logits without insight into the underlying reasoning process. Although this dense supervision is beneficial, it introduces two fundamental limitations:
\textbf{(1) Implicit and uncontrollable supervision}. 
The privileged target is a continuous token-level logit distribution generated through stochastic decoding, whose content, structure, and pedagogical quality cannot be directed. There is no mechanism to control what reasoning the target demonstrates or how it addresses the model's specific errors. The supervision tells the model what its probability distribution should look like at each position, but operates as a black box that provides no explicit insight into the underlying reasoning process. Moreover, the KL objective contains no structured representation of the reasoning transition from incorrect to correct. The model therefore learns what the right distribution looks like, but never practices diagnosing why a particular reasoning step went wrong or how to correct it.
\textbf{(2) Errors are suppressed rather than leveraged}. 
The privileged target is conditional on access to the correct solution, creating a fundamental gap with the model's unconditional generation distribution. When the model diverges into erroneous reasoning, the KL objective penalizes its error distribution through logit-level pressure, forcing it back toward the privileged trajectory. This has two consequences. First, the model learns to avoid error states but is never trained on how to recover when it finds itself in one. The erroneous prefix, which contains the model's own reasoning up to the point of failure, is treated as something to be overwritten rather than as a valuable starting point for learning. Second, the large gradient signals at error positions compress the model's predictive diversity, suppressing the distributional diversity needed for effective exploration and autonomous self-correction. Rather than developing intrinsic error-correction capability, the model becomes increasingly constrained to reproduce a single privileged trajectory.
These limitations point to a deeper issue. Implicit distributional alignment prescribes what the model's distribution should look like without providing explicit reasoning about how to get there. Educational psychology offers a useful lens. The Zone of Proximal Development (ZPD) framework~\citep{vygotsky1978mind} holds that effective instruction must be anchored in the learner's current understanding and must provide an explicit pathway from where the learner is, including their mistakes, to where they need to be.

In this paper, we propose \textbf{T}rajectory-\textbf{A}ugmented \textbf{P}olicy \textbf{O}ptimization (\textbf{\method}), a framework that advances self-distillation from implicit distributional alignment to explicit trajectory construction. \method constructs new training trajectories containing what we term \textbf{Micro-Reflective Corrections}, which are targeted, natural-language transitions in which the model identifies a specific error in its own reasoning, diagnoses its cause, and executes the correction. Each trajectory begins from the model's own erroneous prefix, treating the mistake as the starting point for learning rather than something to be suppressed. The key insight is that each GRPO iteration naturally produces the raw material for this construction, since sampling $K$ solutions per problem reveals precisely where the model can and cannot reason correctly. \method uses this contrastive information to synthesize corrective trajectories anchored in the model's actual error patterns, without requiring privileged conditioning or a separate supervisor. Since each trajectory is rooted in the model's own erroneous prefix and guided by its own correct solutions, the constructed signal stays within the model's distributional neighborhood, making the training targets directly relevant to the model's current reasoning patterns.
To ensure that these trajectories can be effectively integrated into advantage-based RL, \method further introduces \textbf{Difficulty-aware Candidate Selection (DCS)} that targets problems within the model's Zone of Proximal Development and produces an emergent curriculum as the model improves, \textbf{Decoupled Advantage Estimation (DAE)} that prevents the inflated group mean from distorting the base GRPO update, and \textbf{OOD Token Suppression (OTS)} that down-weights out-of-distribution corrective tokens to maintain stable optimization throughout training. All of these mechanisms operate within the standard RL framework, preserving the exploration-exploitation balance without introducing auxiliary KL objectives.

A crucial design choice in \method is that micro-reflective trajectory construction is applied \emph{only during training} to provide corrective supervision. At inference, \method performs standard single-pass generation without thinking mode, just like the base GRPO model. The goal is not to teach the model to produce surface-level reflection patterns at test time, but to internalize the capability of diagnosing and recovering from errors into its general reasoning behavior, enabling the model to autonomously self-correct when necessary, without explicit reflection prompts or multi-turn decoding.

We evaluate \method on three challenging mathematical competition benchmarks (AIME 2024, AIME 2025, and HMMT 2025) and demonstrate consistent improvements over both GRPO and distributional alignment baselines under the same number of training steps. Further analysis through the Direct Solution Rate (DSR) and Effective Reflection Rate (ERR) confirms that \method strengthens both first-pass reasoning and error-correction effectiveness.
The contributions are summarized as follows:

(1) We propose \method, a self-distillation framework that constructs learnable trajectories of micro-reflective corrections from the model's own errors, achieving consistent improvements over GRPO under the same number of training steps.

(2) We identify and resolve key challenges in integrating constructed trajectories into advantage-based RL. Difficulty-aware Candidate Selection targets problems within the model's capability boundary and yields an emergent curriculum. Decoupled Advantage Estimation prevents gradient contamination from inflated group rewards. OOD Token Suppression maintains stable optimization when incorporating out-of-distribution corrective tokens.

(3) Through DSR and ERR analysis, we show that \method strengthens both first-pass reasoning and error-correction effectiveness, confirming that the micro-reflective training signal transfers to the model's general reasoning capability rather than merely teaching surface-level reflection behavior.

%% file: sections/20-related.tex
\section{Related Work}

\subsection{Reinforcement Learning for Reasoning}

Reinforcement learning with verifiable rewards (RLVR) has become a central paradigm for improving the reasoning capabilities of large language models. Early work on STaR~\citep{zelikman2022star} and ReST~\citep{gulcehre2023rest} demonstrated that models can bootstrap their reasoning through iterative self-training on self-generated correct solutions, a principle that underlies much of modern RLVR. DeepSeek-R1~\citep{deepseek-r1} subsequently showed that RL training can elicit sophisticated chain-of-thought reasoning, including self-verification and backtracking behaviors. The GRPO algorithm~\citep{deepseekmath} simplifies policy optimization by using group-relative advantages computed from multiple sampled solutions, eliminating the need for a separate critic model. DAPO~\citep{dapo} further improves training stability through token-level policy gradient loss, dynamic sampling, and clip-higher mechanisms. These methods share a common structure: at each iteration, the model samples multiple candidate solutions from its current policy and receives sparse, outcome-level rewards. The contrastive information within each sampling group, specifically which solutions are correct and which are incorrect, is used only implicitly through advantage normalization.
In this paper, \method builds upon the GRPO framework by making explicit use of this contrastive information. A key observation is that each GRPO iteration naturally produces both correct and incorrect solutions to the same problem, revealing the model's specific error patterns. Rather than treating incorrect solutions as data to be discarded or penalized, \method transforms them into the starting point for constructing corrective trajectories. \method converts the implicit contrastive signal of GRPO groups into explicit, learnable supervision that teaches the model to diagnose and recover from its own errors.

\subsection{Self-Distillation and Error-Driven Learning}

A parallel line of work explores how models can learn from their own outputs through distillation and error-driven correction. On-Policy Distillation (OPD)~\citep{agarwal2024opd} provides dense token-level supervision by having a teacher model guide the student on its own rollouts through KL divergence minimization. On-Policy Self-Distillation (OPSD)~\citep{zhao2026opsd} eliminates the external teacher by using the same model conditioned on privileged information as the teacher. ROSD~\citep{rosd} further improves OPSD by introducing error-focused reflection to guide the teacher's supervision. These methods share a common mechanism: at each token position, the student minimizes KL divergence toward a target distribution that is \emph{external} to the student's current reasoning, whether from a separate teacher or from the same model conditioned on privileged correct information. The supervision is continuous and position-wise, providing no structured representation of the error-to-correction transition. The student learns what the right distribution looks like, but never practices diagnosing why a particular step went wrong or how to recover from it.

The idea of models improving their own outputs through iterative revision has also been explored. Self-Refine~\citep{madaan2023self} prompts a model to critique and revise its own generation in a multi-turn loop. Reflexion~\citep{shinn2023reflexion} uses verbal reinforcement by converting environment feedback into textual reflection signals. SCoRe~\citep{kumar2024training} trains models to self-correct via multi-turn RL, where the model first generates a response and then produces a corrected version. Experiential Reinforcement Learning (ERL)~\citep{erl} embeds an experience-reflection-consolidation loop into RL training for language agents. These approaches operate through multi-turn loops or separate reflection-distillation stages, incurring additional generation cost at deployment, and methods like ERL rely on rich textual feedback from the environment, making them well-suited to agentic tasks but not to mathematical reasoning where only binary correctness signals are available.
The principle that learning can be enhanced by analyzing errors has deep roots in educational psychology~\citep{vygotsky1978mind, metcalfe2017learning}. In LLM training, rejection sampling~\citep{yuan2023scaling} filters generated responses to retain only high-quality outputs, while DPO~\citep{rafailov2023direct} and its variants use preference pairs to train models. These approaches operate by \emph{selecting} or \emph{ranking} existing samples, discarding the structural information about why incorrect solutions failed and providing no mechanism to teach the model how to transition from incorrect to correct reasoning.
In contrast, \method constructs entirely new training trajectories that capture the transition from incorrect to correct reasoning. These trajectories begin from the model's own erroneous prefix, explicitly encode the error-to-correction transition through natural-language analysis, and optimize through advantage-based policy gradients rather than distributional alignment. Crucially, the reflective trajectory construction is applied \emph{only during training} to provide corrective supervision, while at inference \method performs standard single-pass generation without thinking mode, just like the base GRPO model. The constructed trajectories are integrated into single-pass RL training alongside standard rollouts, requiring no multi-turn architecture at training or inference time. The corrective signal is derived from the contrastive structure already present within GRPO sampling groups, making \method applicable to any setting where GRPO is used without additional infrastructure.

\subsection{Training Stability and Distributional Alignment}

Training on data that deviates from the model's own distribution can lead to instability and performance degradation. Distribution Discriminant Theory (DDT)~\citep{ddt} formalizes this insight via a centered log-likelihood criterion that quantifies alignment between a token and the model's predictive distribution, providing a principled measure of in-distribution versus out-of-distribution data. In the RL context, off-policy data introduces bias that requires correction through importance sampling or constrained optimization.
DDT and related distribution-aware methods operate at the sample or trajectory level: they measure whether an entire sequence is in-distribution and decide whether to include or exclude it. This coarse-grained filtering can be overly conservative, discarding trajectories that contain valuable learning signal alongside a few out-of-distribution tokens. \method's OOD Token Suppression (OTS) mechanism draws inspiration from DDT but operates at the \emph{token level}, selectively attenuating out-of-distribution tokens while preserving the learning signal from in-distribution tokens. This fine-grained control is essential for \method, since constructed trajectories inherently contain a mix of in-distribution reasoning (the preserved prefix) and potentially out-of-distribution tokens (the corrective transition), and discarding the entire trajectory would lose the valuable prefix signal.

%% file: sections/40-methods.tex
\section{Method}

\begin{figure}[t]
\centering
\vspace{-10mm}
\includegraphics[width=\textwidth]{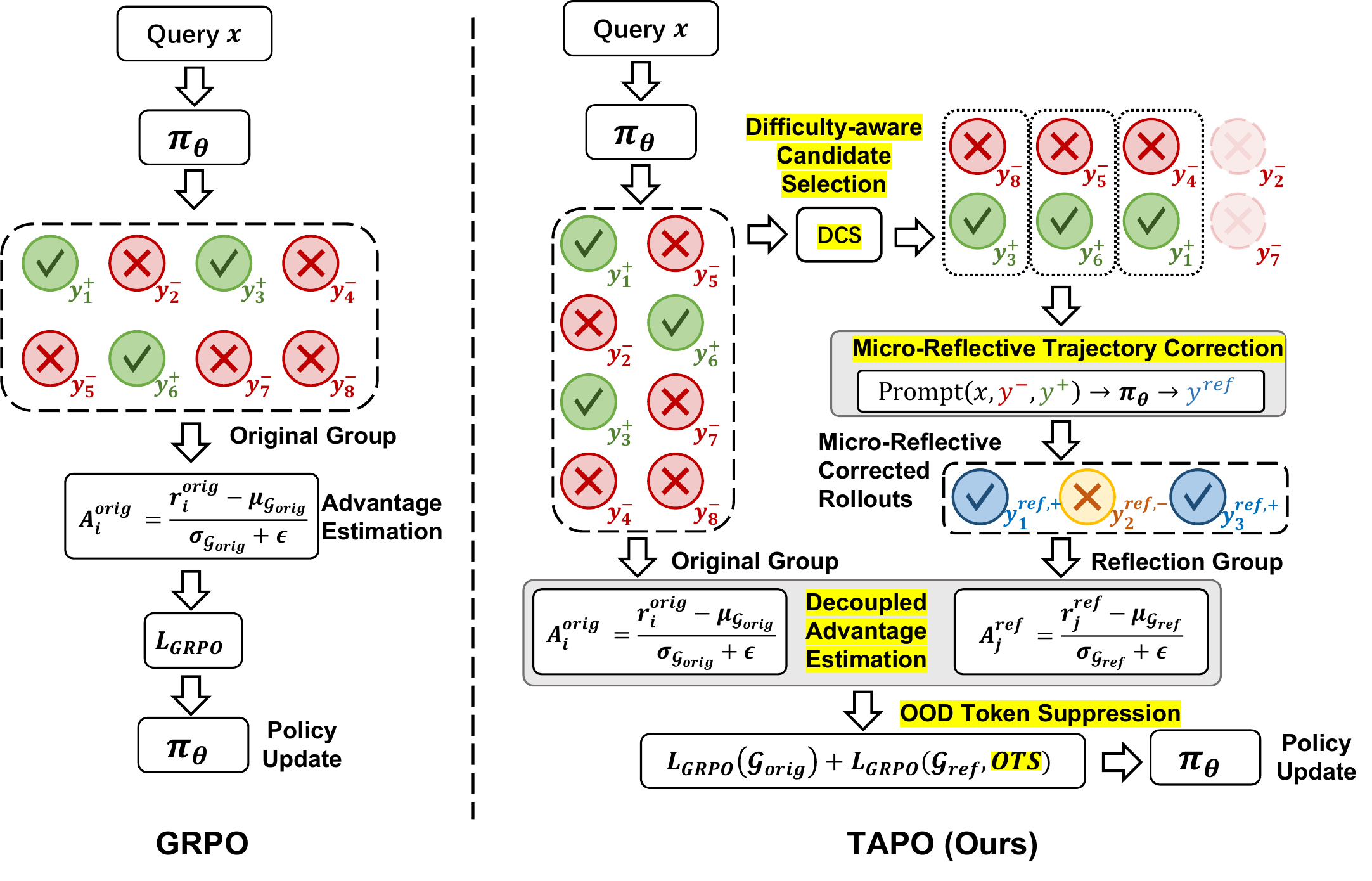}
\caption{Comparison of GRPO and \method. In standard GRPO (top), a single group-relative advantage is computed over all \(K\) rollouts. In \method (bottom), two innovations are introduced. First, Micro-Reflective Trajectory Construction pairs an incorrect rollout \(\mathbf{y}^-\) with a correct reference \(\mathbf{y}^+\) for ZPD problems and synthesizes a micro-reflective trajectory \(\mathbf{y}^{\text{ref}}\) via the actor itself. Second, Decoupled Advantage Estimation normalizes original rollouts and constructed trajectories in independent groups \(\mathcal{G}_{\text{orig}}\) and \(\mathcal{G}_{\text{ref}}\), preventing advantage contamination. The final loss combines both groups with OOD Token Suppression applied to the reflection term.}
\label{fig:overview}
\end{figure}

In this section, we present the Trajectory-Augmented Policy Optimization (\method) framework. The core idea is to construct learnable corrective trajectories that are anchored in the model's own errors and remain within its distributional neighborhood, providing a pedagogically grounded alternative to forced distributional alignment. We begin with the necessary background on GRPO~\citep{deepseekmath} (\S\ref{sec:background}), then describe the cold-start phase (\S\ref{sec:cold_start}) and the two core components of \method, namely Micro-Reflective Trajectory Construction (\S\ref{sec:reflection}) and Decoupled Advantage Estimation (\S\ref{sec:decoupled_adv}), followed by the training stabilization mechanisms used to maintain stable optimization (\S\ref{sec:stabilization}). An overview of the complete \method pipeline is provided in Figure~\ref{fig:overview}.

\subsection{Background on Group Relative Policy Optimization}
\label{sec:background}

Given a query $\mathbf{x}$, GRPO~\citep{deepseekmath} samples a group of $K$ responses $\{\mathbf{y}_1, \ldots, \mathbf{y}_K\}$ from the current policy $\pi_\theta$, and evaluates each response against a verifiable reward function to obtain rewards $\{r_1, \ldots, r_K\}$. GRPO builds on the clipped surrogate objective of PPO~\citep{schulman2017ppo}, replacing the learned value function with a group-relative baseline. The advantage for each response is computed using group-relative normalization as
\begin{equation}
    A_i = \frac{r_i - \mu_{\mathcal{G}}}{\sigma_{\mathcal{G}} + \epsilon}, \quad
    \mu_{\mathcal{G}} = \frac{1}{K}\sum_{j=1}^{K} r_j, \quad
    \sigma_{\mathcal{G}} = \sqrt{\frac{1}{K}\sum_{j=1}^{K}(r_j - \mu_{\mathcal{G}})^2}
    \label{eq:grpo_adv}
\end{equation}
where $\mathcal{G}$ denotes the group of responses sharing the same query. The policy gradient loss with clipped importance sampling is defined as
\begin{equation}
    \mathcal{L}_{\text{GRPO}} = -\frac{1}{T}\sum_{t=1}^{T} \min\left(\rho_t \cdot A_i,\; \text{clip}(\rho_t, 1-\epsilon, 1+\epsilon) \cdot A_i\right)
    \label{eq:grpo_loss}
\end{equation}
where $\rho_t = \frac{\pi_\theta(y_t | \mathbf{x}, \mathbf{y}_{<t})}{\pi_{\theta_{\text{old}}}(y_t | \mathbf{x}, \mathbf{y}_{<t})}$ is the token-level importance sampling ratio.

\subsection{Cold-Start for Trajectory Construction Capability}
\label{sec:cold_start}

Before RL training, we perform a cold-start phase to equip the model with two essential capabilities. First, the model must be able to follow the structured trajectory construction prompt, which requires producing output with specific XML-style tags (\texttt{<analysis>} and \texttt{<reconstruction>}) that a standard pre-trained model does not reliably generate. Second, and more fundamentally, the model must develop a basic ability to analyze its own errors and produce corrections, which serves as the foundation for the micro-reflective trajectory construction during RL training. Without this capability, the policy starts from an immature state where constructed trajectories contain severe OOD tokens, and the OTS mechanism is forced to suppress a large proportion of the corrective signal, as we will quantitatively demonstrate through OTS weight dynamics analysis in \S\ref{sec:training_dynamics}.

The cold-start phase constructs a total of 45,000 training examples using the full set of approximately 40,000 queries from the DeepScaleR~\citep{deepscaler} dataset. For each query, we sample responses from the base model Qwen3-8B-Instruct~\citep{qwen3}, partition them into correct and incorrect groups, and then construct micro-reflective corrective trajectories from the incorrect responses, with at most one corrective trajectory per query. This yields a joint training set consisting of 30,000 SFT-format examples and 15,000 IFT-format examples, trained jointly to equip the model with both trajectory construction and instruction-following capabilities. The cold-start model serves as the initialization for all subsequent RL experiments.

\subsection{Micro-Reflective Trajectory Construction}
\label{sec:reflection}

The core innovation of \method is the \textbf{Micro-Reflective Trajectory Construction} process, which transforms the contrastive information in GRPO groups into learnable training signal. This process operates within each GRPO training iteration and consists of three steps, namely candidate selection, trajectory synthesis, and reward evaluation.

\subsubsection{Difficulty-Aware Candidate Selection}
\label{sec:candidate_selection}

For each query $\mathbf{x}$ in the current training batch, we examine the $K$ sampled responses and partition them into correct ($\mathcal{P} = \{i : r_i > 0\}$) and incorrect ($\mathcal{N} = \{i : r_i = 0\}$) sets based on their accuracy reward. A query is eligible for reflective reconstruction if and only if the following condition holds.
\begin{equation}
    |\mathcal{P}| \geq n_{\text{pos}} \quad \text{and} \quad |\mathcal{N}| \geq n_{\text{neg}}
    \label{eq:zpd_condition}
\end{equation}
where $n_{\text{pos}}$ and $n_{\text{neg}}$ are hyperparameters that define the Zone of Proximal Development (ZPD) boundary~\citep{vygotsky1978mind}. This condition naturally partitions problems into three difficulty regions relative to the model's current capability: the mastered zone where the model almost always succeeds, the ZPD where the model has genuine but inconsistent understanding, and the beyond-capability zone where the model rarely succeeds. Trajectory construction is only applied to ZPD queries. As training progresses, the ZPD boundary shifts dynamically, creating an emergent curriculum~\citep{bengio2009curriculum} where the model is continually challenged with problems at the frontier of its capability. For each eligible query, we select up to $m_{\text{max}}$ incorrect responses from $\mathcal{N}$, and for each selected incorrect response $\mathbf{y}^-$, we randomly sample a correct reference $\mathbf{y}^+ \in \mathcal{P}$ to form a construction pair $(\mathbf{y}^-, \mathbf{y}^+)$. This yields up to $m_{\text{max}}$ correctable samples per eligible query.

\subsubsection{Trajectory Synthesis}
\label{sec:reflection_generation}

For each selected construction pair $(\mathbf{y}^-, \mathbf{y}^+)$, we construct a synthesis prompt that presents the original question, the incorrect response, and the correct reference, and instructs the model to perform two operations. First, the model \textbf{analyzes} the error by identifying the location and type of the first critical mistake in the incorrect response. Second, the model \textbf{constructs} a corrective trajectory by reproducing the incorrect reasoning up to and including the error point, inserting a natural correction transition, and then continuing with correct reasoning to reach the final answer.

The construction process, which involves explicit error analysis and correction, can produce tokens that deviate from the model's natural generation distribution. We refer to the resulting output as a \textbf{micro-reflective trajectory}, reflecting its granular nature. It preserves the learner's valid reasoning prefix and only intervenes from the first error point onward.

\begin{figure}[t]
\centering
\vspace{-10mm}
\includegraphics[width=\textwidth]{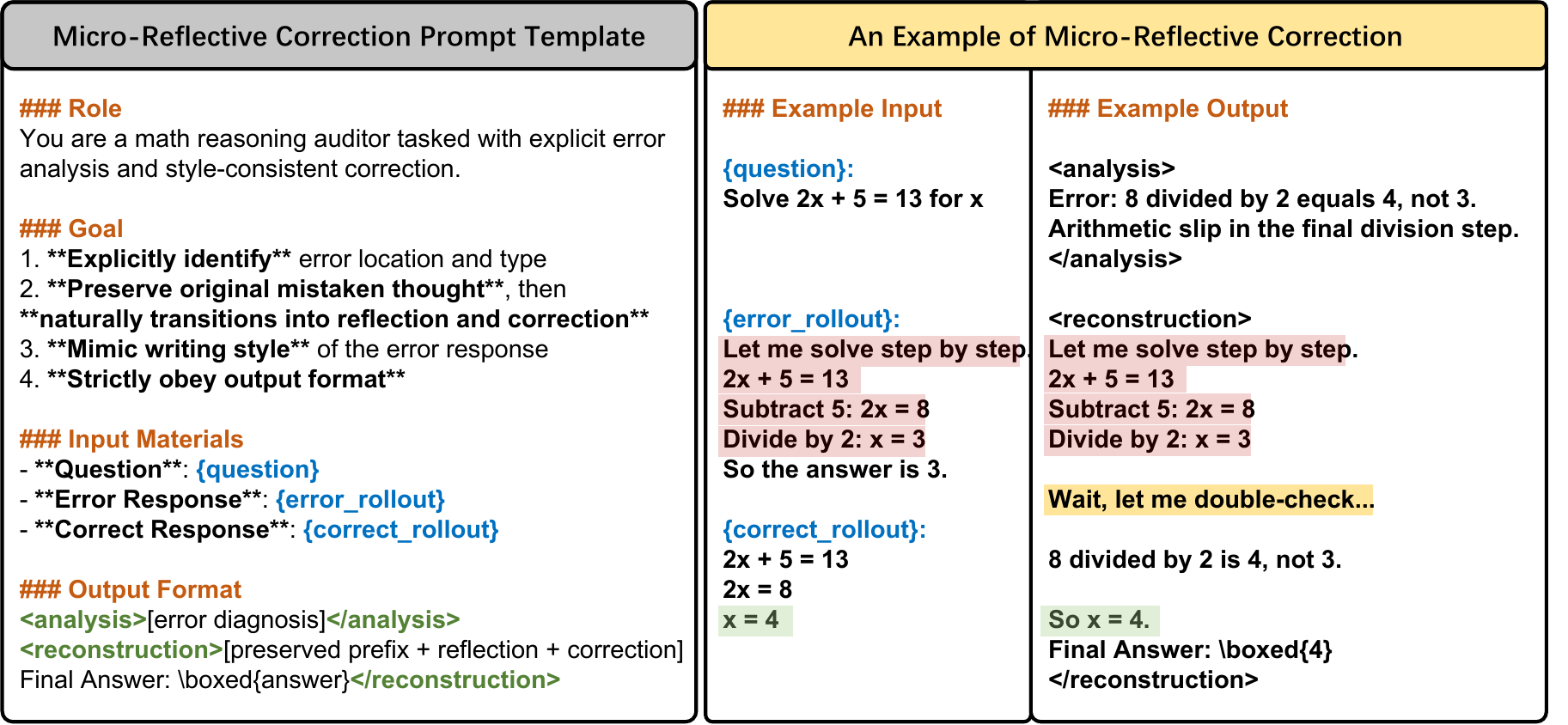}
\caption{Illustration of the Micro-Reflective Trajectory Construction prompt template. The left panel shows the synthesis prompt instructing the model to analyze the error and construct a corrective trajectory that preserves the incorrect reasoning verbatim up to the first mistake, inserts a natural reflection phrase, and continues with corrected reasoning. The middle panel presents a concrete example input with the original question, an error response containing an arithmetic slip at the final step, and the correct reference solution. The right panel shows the resulting micro-reflective trajectory, where the error prefix is preserved verbatim, a natural reflection phrase triggers the correction, and the corrected reasoning continues to the final answer. Red denotes the error prefix, yellow the reflection phrase, and green the corrected reasoning.}
\vspace{-5mm}
\label{fig:micro_reflective}
\end{figure}

Figure~\ref{fig:micro_reflective} illustrates the micro-reflective construction process with a concrete example. The key design principle is that the error prefix is preserved exactly as generated by the model, maintaining distributional proximity, while the reflection phrase serves as a natural cognitive transition that the model can learn to internalize. This contrasts with full reconstruction approaches that discard the entire incorrect reasoning and regenerate from scratch, which we show in \S\ref{sec:reflective_ablation} leads to degraded performance due to distributional mismatch.

It is important to note that micro-reflective trajectory construction is applied \textbf{only during training}. At inference, \method performs standard single-pass generation without thinking mode, identical to the base GRPO model. The training-time construction introduces no additional inference cost, and the goal is to internalize error-correction capability into the model's general reasoning behavior rather than teaching it to produce explicit reflection patterns at test time.

\subsection{Decoupled Advantage Estimation}
\label{sec:decoupled_adv}

A straightforward approach to incorporating constructed trajectories would be to append them to the original GRPO group and compute advantages jointly. However, this leads to a phenomenon we term \textbf{advantage contamination}.

Consider a query $\mathbf{x}$ with original group $\mathcal{G}_{\text{orig}} = \{\mathbf{y}_1, \ldots, \mathbf{y}_K\}$ containing $|\mathcal{P}|$ correct and $|\mathcal{N}|$ incorrect responses, with rewards $r^+ > 0$ and $r^- = 0$ respectively. The original group mean reward is $\mu_{\mathcal{G}_{\text{orig}}} = |\mathcal{P}| \cdot r^+ / K$. If we add $m$ constructed trajectory samples, of which $m^+$ are correct (reward $r^+$) and $m^- = m - m^+$ are incorrect, the contaminated mean becomes $\mu_{\text{mixed}} = \frac{(|\mathcal{P}| + m^+) \cdot r^+}{K + m} > \mu_{\mathcal{G}_{\text{orig}}}$ when $m^+ > 0$. This elevated mean causes the advantages of the original incorrect samples to become more negative, as $A^-_{\text{mixed}} = (0 - \mu_{\text{mixed}}) / \sigma_{\text{mixed}} < (0 - \mu_{\mathcal{G}_{\text{orig}}}) / \sigma_{\mathcal{G}_{\text{orig}}} = A^-_{\text{orig}}$. Since incorrect samples are typically longer than correct ones (due to more extensive but ultimately flawed reasoning chains), this amplified negative advantage indirectly creates a penalty on response length, leading to three cascading failures, namely response length collapse, policy entropy collapse, and ultimately performance degradation as exploration capacity diminishes.

To resolve this, we compute advantages for original and constructed trajectory samples independently. We assign each constructed trajectory a distinct group identifier by appending a suffix to the original query's uid. For a query with uid $\text{id}$, the original $K$ rollouts retain uid $\text{id}$, while the constructed trajectories receive uid $\text{id\_reflected}$. Since GRPO advantage normalization operates per uid, the original and reflection groups are automatically treated as separate groups. The reflection group $\mathcal{G}_{\text{ref}}$ consists of all constructed trajectories for the query. Among these, those that produce the correct answer serve as positive samples ($r^{\text{ref}} = 1$), while those that fail to reach the correct answer serve as negative samples ($r^{\text{ref}} = 0$). With $m_{\text{max}} = 4$ and a parsing success rate of 80~90\%, each eligible query contributes approximately 3 to 4 constructed trajectories to $\mathcal{G}_{\text{ref}}$. The original group $\mathcal{G}_{\text{orig}}$ retains the $K$ rollouts unchanged. Advantages are computed independently within each group.
\begin{align}
    A^{\text{orig}}_i &= \frac{r^{\text{orig}}_i - \mu_{\mathcal{G}_{\text{orig}}}}{\sigma_{\mathcal{G}_{\text{orig}}} + \epsilon}, \quad \text{for } \mathbf{y}_i \in \mathcal{G}_{\text{orig}} \\
    A^{\text{ref}}_j &= \frac{r^{\text{ref}}_j - \mu_{\mathcal{G}_{\text{ref}}}}{\sigma_{\mathcal{G}_{\text{ref}}} + \epsilon}, \quad \text{for } \mathbf{y}^{\text{ref}}_j \in \mathcal{G}_{\text{ref}}
    \label{eq:decoupled_adv}
\end{align}

This decoupling ensures that the reflective trajectory construction provides \textbf{purely additive} gradient signal without distorting the base GRPO update for original samples. Within the reflection group, the contrastive structure of GRPO is preserved. Correct trajectories receive positive advantages and incorrect ones receive negative advantages, allowing the model to learn to distinguish effective corrections from ineffective ones. When $|\mathcal{G}_{\text{ref}}| = 1$, group normalization is undefined, so we directly use the reward as the advantage ($A^{\text{ref}} = r^{\text{ref}}$), equivalent to weighted SFT on this single sample. The final training loss combines both groups as
\begin{equation}
    \mathcal{L}_{\text{\method}} = \mathcal{L}_{\text{GRPO}}(\mathcal{G}_{\text{orig}}) + \lambda \cdot \mathcal{L}_{\text{ref}}(\mathcal{G}_{\text{ref}})
    \label{eq:tapo_loss}
\end{equation}
where $\mathcal{L}_{\text{GRPO}}$ is the standard GRPO loss on the original group as defined in Eq.~\ref{eq:grpo_loss}, $\mathcal{L}_{\text{ref}}$ employs the same clipped surrogate objective on the reflection group but with the decoupled advantage $A^{\text{ref}}_j$ from Eq.~\ref{eq:decoupled_adv}, and $\lambda$ controls the relative weight of the reflection trajectory loss, set to 1.0 in all experiments. The detailed formulation of $\mathcal{L}_{\text{ref}}$ is provided in \S\ref{sec:stabilization} (Eq.~\ref{eq:ref_loss}).

\subsection{Training Stabilization}
\label{sec:stabilization}

Although constructed trajectories originate from the model itself, the construction process may produce tokens that deviate from the model's natural generation distribution, particularly at corrective transition points. To prevent these out-of-distribution tokens from destabilizing training, we apply a token-level weighting mechanism that we term \textbf{OOD Token Suppression (OTS)} to the reflection trajectory loss, drawing inspiration from Distribution Discriminant Theory (DDT)~\citep{ddt}.

For each token $y_t$ in a constructed trajectory, we compute a distribution alignment score that measures how well it aligns with the model's current distribution as
\begin{equation}
    s_t = \log p_\theta(y_t | \mathbf{x}, \mathbf{y}_{<t}) + H[p_\theta(\cdot | \mathbf{x}, \mathbf{y}_{<t})]
    \label{eq:alignment_score}
\end{equation}
where $H[p_\theta]$ is the entropy of the predicted distribution at that position. The score is converted into a multiplicative weight $w_t = \text{clamp}(\exp(s_t), w_{\min}, w_{\max})$. Tokens that are natural for the model receive weights near 1.0, while out-of-distribution corrective tokens are automatically down-weighted. This selective attenuation preserves the learning signal from mathematical reasoning portions while reducing the gradient contribution of meta-cognitive transition phrases.

The reflection trajectory loss applies this OTS weighting as
\begin{equation}
    \mathcal{L}_{\text{ref}} = -\frac{1}{T}\sum_{t=1}^{T} w_t \cdot \min\left(\rho_t \cdot A^{\text{ref}},\; \text{clip}(\rho_t, 1-\epsilon, 1+\epsilon) \cdot A^{\text{ref}}\right)
    \label{eq:ref_loss}
\end{equation}
The weighting is applied exclusively to the constructed trajectory samples. Original GRPO samples use the standard unweighted loss, ensuring that the base training signal remains undistorted.
And the complete \method training procedure for one iteration is summarized in Algorithm~\ref{alg:tapo}.

\begin{algorithm}[h]
\caption{Trajectory-Augmented Policy Optimization (\method), One Training Iteration}
\label{alg:tapo}
\begin{algorithmic}[1]
\Require Policy $\pi_\theta$, batch of queries $\mathcal{B}$, group size $K$, ZPD thresholds $n_{\text{pos}}, n_{\text{neg}}$, max constructions $m_{\text{max}}$
\State \textbf{// Standard GRPO Rollout}
\For{each query $\mathbf{x} \in \mathcal{B}$}
    \State Sample $K$ responses $\{\mathbf{y}_1, \ldots, \mathbf{y}_K\} \sim \pi_\theta(\cdot | \mathbf{x})$
    \State Evaluate rewards $\{r_1, \ldots, r_K\}$ via verifiable reward function
    \State Partition into correct $\mathcal{P}$ and incorrect $\mathcal{N}$ based on accuracy reward
\EndFor
\State \textbf{// Micro-Reflective Trajectory Construction}
\For{each query $\mathbf{x}$ where $|\mathcal{P}| \geq n_{\text{pos}}$ and $|\mathcal{N}| \geq n_{\text{neg}}$}
    \State Select up to $m_{\text{max}}$ incorrect responses $\mathcal{N}_{\text{sel}} \subseteq \mathcal{N}$
    \For{each $\mathbf{y}^- \in \mathcal{N}_{\text{sel}}$}
        \State Sample reference $\mathbf{y}^+ \sim \text{Uniform}(\mathcal{P})$
        \State Construct synthesis prompt from $(\mathbf{x}, \mathbf{y}^-, \mathbf{y}^+)$
        \State Generate corrective trajectory $\mathbf{y}^{\text{ref}} \sim \pi_\theta(\cdot | \text{prompt})$ via vLLM
        \State Parse and evaluate, obtain $r^{\text{ref}}$ from reward function
    \EndFor
\EndFor
\State \textbf{// Decoupled Advantage Estimation}
\State Compute $A^{\text{orig}}_i$ for original samples using Eq.~\ref{eq:grpo_adv} within $\mathcal{G}_{\text{orig}}$
\State Compute $A^{\text{ref}}_j$ for constructed trajectories using Eq.~\ref{eq:decoupled_adv} within $\mathcal{G}_{\text{ref}}$
\State \textbf{// Training Stabilization}
\State Compute token-level weights $w_t$ for constructed trajectories using Eq.~\ref{eq:alignment_score}
\State Update $\theta$ using combined loss $\mathcal{L}_{\text{\method}} = \mathcal{L}_{\text{GRPO}}(\mathcal{G}_{\text{orig}}) + \mathcal{L}_{\text{ref}}(\mathcal{G}_{\text{ref}})$ (Eq.~\ref{eq:tapo_loss}, \ref{eq:ref_loss})
\end{algorithmic}
\end{algorithm}

The primary computational overhead of \method relative to standard GRPO comes from the trajectory construction step, which requires an additional forward pass through the model for each selected construction pair. In practice, with $m_{\text{max}} = 4$ and typical ZPD eligibility rates, the trajectory construction triggers on a modest subset of problems per batch, and the synthesis is parallelized through vLLM batch inference. \method incurs zero additional cost at inference time. The trained model is deployed with standard single-pass inference.

%% file: sections/50-experiments.tex
\section{Experiments}
\label{sec:experiments}

We evaluate \method on challenging mathematical reasoning benchmarks and organize our experiments around the central questions of this work. We first present overall results comparing \method with GRPO and the self-distillation baseline (\S\ref{sec:main_results}). We then examine whether the observed improvements reflect genuine capability internalization in both first-pass reasoning and error correction (\S\ref{sec:internalization}). Next, we investigate corrective trajectory learning through controlled ablations of trajectory structure, construction granularity, and construction source (\S\ref{sec:reflective_ablation}), and analyze the distributional proximity mechanism that makes self-constructed micro-reflective corrections effective for training. Finally, we verify the necessity of each integration component through ablation studies and training dynamics analysis (\S\ref{sec:integration}).

\subsection{Experimental Setup}
\label{sec:setup}

\paragraph{Training Data and Base Model.} We use Qwen3-8B-Instruct~\citep{qwen3} as the base model. Our training queries are drawn from the DeepScaleR~\citep{deepscaler} dataset, which consists of approximately 40,000 unique mathematics problem-answer pairs compiled from AIME 1984-2023, AMC (prior to 2023), Omni-Math, and STILL. To ensure that the micro-reflective trajectory construction can be effectively triggered during RL training, we filter queries based on the cold-start model's accuracy on DeepScaleR, retaining only those with accuracy in the range of 0.125 to 0.875. This yields approximately 16,000 training queries after filtering. Queries with accuracy below 0.125 are too difficult for the model to produce any correct reference, while those above 0.875 are too easy to generate insufficient incorrect samples, both of which prevent the ZPD-based candidate selection from activating. Each query is used once per epoch. Prior to RL training, we perform a cold-start supervised fine-tuning phase to equip the model with basic trajectory construction capability (\S\ref{sec:cold_start}). The cold-start phase uses a learning rate of 5e-6 with linear warmup over 50 steps and cosine decay, trained jointly on 30,000 SFT and 15,000 IFT examples for 3 epochs. The cold-start model serves as the initialization for all subsequent RL experiments.

\paragraph{Training Configuration.} We train with GRPO using a group size of $K=8$. The RL optimizer is AdamW with a learning rate of 1e-6, constant warmup, and cosine decay schedule. For \method, we set the ZPD thresholds to $n_{\text{pos}}=2$ and $n_{\text{neg}}=4$, with maximum constructions $m_{\text{max}}=4$ per eligible query. Token-level weight bounds are $w_{\min}=0.01$ and $w_{\max}=10.0$. All models are trained for one epoch with 32 batch size, i.e., approximately 500 steps, unless otherwise noted. OPSD~\citep{zhao2026opsd} uses the same training configuration as GRPO and \method to ensure fair comparison. 

\paragraph{Evaluation Benchmarks.} We evaluate on three challenging mathematical competition benchmarks. AIME 2024 is the American Invitational Mathematics Examination 2024, consisting of 30 problems across two contests. AIME 2025 is the American Invitational Mathematics Examination 2025, with the same format but updated problems. HMMT 2025 is the Harvard-MIT Mathematics Tournament 2025, featuring competition-level problems.

\paragraph{Evaluation Protocol.} For each benchmark, we report Pass@$k$ for $k \in \{1, 2, 3, 4, 5\}$ by sampling 5 solutions per problem with temperature 0.6 and top-$p$ 0.9. All methods use the non-thinking mode of the base model unless explicitly labeled as ``Thinking''. All reported results are averaged over \textbf{16} independent evaluation runs to ensure statistical reliability. The thinking mode is disabled by default to ensure consistent evaluation conditions across all methods. Importantly, \method performs standard single-pass generation at inference time without any explicit reflection prompts or multi-turn decoding; the micro-reflective trajectory construction is applied only during training to provide corrective supervision, and the goal is to internalize error-correction capability into the model's general reasoning behavior.

\paragraph{Baselines.} We compare against the following methods. \textbf{Qwen3-8B-Instruct} is the base model without any further training. \textbf{Qwen3-8B-Instruct w/ Thinking} uses the same base model with thinking mode enabled at inference time. \textbf{Cold-Start} is the cold-start checkpoint before RL training. \textbf{GRPO} is standard GRPO training without micro-reflective trajectory construction. \textbf{OPSD} applies On-Policy Self-Distillation~\citep{zhao2026opsd}, which uses the model's own log-probabilities on correct solutions as the privileged target distribution. All RL training baselines, namely GRPO, OPSD, and \method, are evaluated in two settings, namely starting from the Qwen3-8B-Instruct base model directly, and starting from the cold-start checkpoint. Both settings train for the same number of steps to ensure fair comparison.

\subsection{Main Results}
\label{sec:main_results}

\begin{table*}[t]
\centering
\caption{Main results across two training settings: Direct training from Qwen3-8B-Instruct in the top half of each benchmark and cold-start initialization in the bottom half. Pass@$k$ ($k=1,\dots,5$) reported for all methods. Bold and underline indicate best and second-best results within each setting.}
\label{tab:main_results}
\resizebox{0.8\textwidth}{!}{
\begin{tabular}{l|ccccc}
\toprule
\textbf{Method} & \textbf{Pass@1} & \textbf{Pass@2} & \textbf{Pass@3} & \textbf{Pass@4} & \textbf{Pass@5} \\
\midrule
\multicolumn{6}{c}{\textbf{AIME 2024}} \\
\midrule
Qwen3-8B-Instruct & 23.12 & 31.52 & 35.00 & 40.00 & 42.71 \\
Qwen3-8B-Instruct w/ Thinking & 55.21 & 62.71 & 66.25 & 68.45 & 70.42 \\
Qwen3-8B-Instruct + OPSD & 54.58 & 63.75 & 69.58 & 72.71 & 75.83 \\
Qwen3-8B-Instruct + GRPO & 54.58 & 63.12 & 66.88 & 69.58 & 72.08 \\
Qwen3-8B-Instruct + \textbf{\method} & \textbf{63.12} & \textbf{71.25} & \underline{74.79} & \underline{76.46} & 77.08 \\
Cold-start & 31.67 & 39.38 & 44.17 & 48.12 & 51.04 \\
Cold-start + OPSD & 57.71 & \underline{65.42} & 70.21 & 75.62 & \underline{78.12} \\
Cold-start + GRPO & 52.92 & 62.92 & 70.00 & 72.50 & 74.58 \\
Cold-start + \textbf{\method} & \underline{62.50} & \textbf{71.25} & \textbf{76.04} & \textbf{77.92} & \textbf{80.21} \\
\midrule
\multicolumn{6}{c}{\textbf{AIME 2025}} \\
\midrule
Qwen3-8B-Instruct & 18.54 & 25.21 & 27.50 & 30.00 & 32.71 \\
Qwen3-8B-Instruct w/ Thinking & 42.08 & 46.67 & 49.38 & 51.46 & 53.12 \\
Qwen3-8B-Instruct + OPSD & 42.29 & 49.58 & 54.37 & 56.67 & 59.58 \\
Qwen3-8B-Instruct + GRPO & 42.08 & 51.04 & 56.88 & 60.42 & 62.50 \\
Qwen3-8B-Instruct + \textbf{\method} & 41.25 & 49.58 & 54.37 & 57.50 & 59.17 \\
Cold-start & 23.75 & 28.96 & 32.50 & 35.42 & 37.08 \\
Cold-start + OPSD & \underline{43.33} & 53.75 & 59.58 & 62.29 & 64.17 \\
Cold-start + GRPO & \textbf{46.88} & \textbf{57.08} & \underline{60.42} & \underline{63.12} & \underline{64.79} \\
Cold-start + \textbf{\method} & \textbf{46.88} & \underline{56.46} & \textbf{61.46} & \textbf{66.04} & \textbf{67.92} \\
\midrule
\multicolumn{6}{c}{\textbf{HMMT 2025}} \\
\midrule
Qwen3-8B-Instruct & 9.79 & 14.17 & 17.50 & 19.38 & 20.83 \\
Qwen3-8B-Instruct w/ Thinking & 25.62 & 30.62 & 34.17 & 36.88 & 38.33 \\
Qwen3-8B-Instruct + OPSD & 22.08 & 28.33 & 33.33 & 35.83 & 37.29 \\
Qwen3-8B-Instruct + GRPO & 25.00 & 29.17 & 32.50 & 35.62 & 37.08 \\
Qwen3-8B-Instruct + \textbf{\method} & 22.71 & 28.54 & 31.87 & 35.21 & 36.88 \\
Cold-start & 14.37 & 18.54 & 21.25 & 23.96 & 26.46 \\
Cold-start + OPSD & 24.17 & 33.33 & 37.92 & 42.08 & 46.46 \\
Cold-start + GRPO & \underline{28.75} & \underline{38.96} & \underline{42.71} & \underline{45.62} & \underline{48.12} \\
Cold-start + \textbf{\method} & \textbf{31.46} & \textbf{40.83} & \textbf{44.79} & \textbf{47.50} & \textbf{50.00} \\
\bottomrule
\end{tabular}
}
\end{table*}

Table~\ref{tab:main_results} presents the main results across two experimental settings. We first discuss the direct training setting, and then analyze the cold-start setting which is the primary configuration for \method.

\textbf{Cold-start Setting.} With cold-start initialization, \method achieves the best Pass@1 on all three benchmarks, outperforming both GRPO and OPSD. On AIME 2024, \method reaches 62.50\% Pass@1, a 9.58-point improvement over Cold-start + GRPO at 52.92\% and a 4.79-point gain over Cold-start + OPSD at 57.71\%. On the more challenging AIME 2025, \method matches GRPO at 46.88\% Pass@1 while outperforming OPSD by 3.55 points. On HMMT 2025, the hardest benchmark, \method achieves 31.46\% Pass@1, surpassing GRPO by 2.71 points and OPSD by 7.29 points. The consistent gains across all three benchmarks, particularly on the hardest setting of HMMT 2025, suggest that \method's explicit trajectory construction is most beneficial when the reasoning task is challenging and the model's initial accuracy is low, providing richer opportunities for learning from errors. The improvements are sustained across all Pass@$k$ values, indicating that the gains reflect genuine capability enhancement rather than variance reduction in a single generation pathway.

\textbf{Direct Training Setting, without Cold-start.} Without cold-start initialization, the results are more nuanced. We first note the strong inference-time baseline.
Qwen3-8B-Instruct w/ Thinking achieves 55.21\% on AIME 2024, 42.08\% on AIME 2025, and 25.62\% on HMMT 2025, substantially outperforming the non-thinking base model at 23.12\%, 18.54\%, and 9.79\%, respectively. Among RL-trained methods, on AIME 2024, \method achieves 63.12\% Pass@1, and 8.54 point improvement over both GRPO and OPSD. However, on AIME 2025 and HMMT 2025, \method underperforms GRPO, with Pass@1 gaps of 0.83 and 2.29 points respectively. This pattern is consistent with the OTS weight dynamics analysis in \S\ref{sec:training_dynamics}.
Without cold start, the policy is not pre-aligned to the trajectory construction format, causing the OTS mechanism to suppress a larger fraction of the corrective signal. The degradation is most visible on harder benchmarks where the model's immature policy produces lower-quality constructions, reducing the effective learning signal. The contrast between the cold-start and direct training settings underscores the importance of the cold-start phase, which provides the policy with sufficient trajectory construction capability before RL training begins. We further analyze the contributions of trajectory structure and RL optimization through controlled ablation in \S\ref{sec:reflective_ablation}.

\subsection{Capability Internalization Analysis}
\label{sec:internalization}

A central claim of \method is that the observed improvements stem from the internalization of autonomous error-correction capability rather than superficial data augmentation. We verify this claim through two complementary metrics that quantify both first-pass reasoning strength and reflective correction effectiveness. Both metrics are computed on the boundary subset of questions where both models exhibit partial success (pass rate strictly between 0 and 1), and are estimated via a stronger judge model (Qwen3-235B-A22B) that classifies each response for the presence, accuracy, and effectiveness of reflective reasoning.

The Direct Solution Rate (DSR) measures the fraction of correctly answered problems where the model reaches the answer directly, without any intermediate reflection or self-correction. A higher DSR indicates stronger first-pass reasoning. Improvements in DSR are particularly informative for \method, because they demonstrate that training on corrective trajectories strengthens the model's underlying reasoning even when no correction takes place. The Effective Reflection Rate (ERR) measures the quality of the model's reflective reasoning when it does engage in self-correction. Among responses that contain reflection and correction attempts, ERR is the fraction that successfully locate the error, provide an accurate diagnosis, and arrive at the correct final answer. DSR captures first-pass reasoning strength, while ERR captures the model's ability to recover when its initial reasoning is incorrect.

\begin{table}[h]
\centering
\caption{Capability internalization: DSR and ERR comparing Cold-start + GRPO and Cold-start + \method. $\Delta$ denotes the absolute improvement of \method over GRPO.}
\label{tab:internalization}
\resizebox{\linewidth}{!}{
\begin{tabular}{l|ccc|ccc}
\toprule
 & \multicolumn{3}{c|}{\textbf{DSR} $\uparrow$} & \multicolumn{3}{c}{\textbf{ERR} $\uparrow$} \\
\textbf{Method} & \textbf{AIME 24} & \textbf{AIME 25} & \textbf{HMMT 25} & \textbf{AIME 24} & \textbf{AIME 25} & \textbf{HMMT 25} \\
\midrule
Cold-start + GRPO & 34.0 & 22.1 & 28.0 & 52.0 & 51.5 & 33.8 \\
Cold-start + \textbf{\method} & 47.5 & 38.0 & 50.4 & 63.4 & 56.3 & 36.8 \\
\midrule
\textbf{$\Delta$ (\method $-$ GRPO)} & +13.5 & +15.9 & +22.3 & +11.4 & +4.8 & +3.0 \\
\bottomrule
\end{tabular}
}
\end{table}

Table~\ref{tab:internalization} reveals a critical finding. \method improves not only the model's error-correction capability but also its first-pass reasoning strength. The DSR of Cold-start + \method is substantially higher than that of Cold-start + GRPO across all benchmarks. This finding is significant because DSR measures correct answers obtained \textit{without any reflection} and the model simply reasons correctly from the start. Training on corrective trajectories improves first-pass reasoning through two hypothesized mechanisms. First, the process of constructing and learning from corrective trajectories forces the model to internalize patterns of common errors. When the model later encounters similar reasoning situations, it implicitly avoids these error patterns, producing correct reasoning on the first attempt. Second, the micro-reflective construction preserves the model's valid reasoning prefix up to the error point. By training on these prefixes paired with corrected continuations, the model learns to extend valid reasoning chains more reliably, effectively strengthening the foundations of its reasoning process. The magnitude of DSR improvement varies across benchmarks. The largest gain occurs on HMMT 2025, followed by AIME 2025 and AIME 2024, suggesting that harder benchmarks with more systematic errors benefit most from corrective training.

The ERR metric provides complementary evidence. \method achieves higher ERR than GRPO across all benchmarks, with the improvement most pronounced on AIME 2024 and more modest on AIME 2025 and HMMT 2025. Collectively, these results support the central claim that \method develops genuine error-correction capability rather than superficial reflection behavior. DSR improvements demonstrate that the training signal transfers to first-pass reasoning, not just reflective responses. ERR improvements show that when reflection occurs, it is more likely to be effective. This indicates that the model has learned to apply reflection selectively and appropriately rather than generating generic meta-cognitive phrases.

\subsection{Ablation of Reflective Trajectory Learning}
\label{sec:reflective_ablation}

This section investigates the source of \method's improvements by disentangling the key design factors behind corrective trajectory learning. All ablation experiments are trained for 200 steps.
Table~\ref{tab:reflective_ablation} reports the ablation results, where each variant makes a single change relative to \method. \textbf{SFT w/ micro-reflective} isolates the contribution of reinforcement learning by training on the same micro-reflective data with maximum likelihood estimation. \textbf{Full reconstruction} removes prefix preservation by replacing the entire solution with a complete regeneration from the model's current policy.

\begin{table*}[h]
\centering
\caption{Ablation of corrective trajectory learning (200 steps). All methods initialized from Cold-start.}
\label{tab:reflective_ablation}
\resizebox{\linewidth}{!}{
\begin{tabular}{l|cc|cc|cc}
\toprule
 & \multicolumn{2}{c|}{\textbf{AIME 2024}} & \multicolumn{2}{c|}{\textbf{AIME 2025}} & \multicolumn{2}{c}{\textbf{HMMT 2025}} \\
\textbf{Configuration} & \textbf{Pass@1} & \textbf{Pass@5} & \textbf{Pass@1} & \textbf{Pass@5} & \textbf{Pass@1} & \textbf{Pass@5} \\
\midrule
Cold-start & 31.67 & 51.04 & 23.75 & 37.08 & 14.37 & 26.46 \\
\midrule
\quad + GRPO & 41.46 & \underline{66.67} & 30.00 & 52.50 & 20.00 & 36.67 \\
\quad + SFT w/ micro-reflective & 30.62 & 44.58 & 20.21 & 32.50 & 12.29 & 21.67 \\
\quad + \method w/ full reconstruction & \underline{42.29} & \textbf{67.08} & \underline{33.75} & \underline{53.96} & 18.75 & \underline{37.50} \\
\quad + \textbf{\method w/ micro-reflective} & \textbf{46.67} & \textbf{67.08} & \textbf{36.46} & \textbf{60.21} & \textbf{20.62} & \textbf{40.21} \\
\bottomrule
\end{tabular}
}
\end{table*}

\paragraph{RL Optimization.} The gap between SFT w/ micro-reflective and \method isolates the contribution of RL optimization. Both use identical micro-reflective trajectory data, but SFT applies uniform maximum likelihood to all tokens, providing no signal about which corrections are effective. \method's advantage-based RL, by contrast, computes group-relative advantages within the reflection group, assigning positive advantages to successful corrections and negative advantages to unsuccessful ones. This contrastive structure enables the model to distinguish effective corrections from ineffective ones, which is a signal that maximum likelihood estimation cannot provide. The results confirm that RL optimization provides a meaningful advantage over supervised imitation even when the trajectory data is identical.

\paragraph{Micro-Reflective Design.} This section isolates the contribution of prefix preservation, a key design choice in micro-reflective trajectory construction. \textbf{\method w/ full reconstruction} removes prefix preservation. Rather than preserving the learner's valid reasoning prefix and only correcting from the error point onward, it replaces the entire solution with a complete regeneration from scratch. The results show that full reconstruction underperforms \method across all benchmarks. On AIME 2024, full reconstruction achieves 42.29\% Pass@1 compared to 46.67\% for micro-reflective, a drop of 4.38 points. The gap is similarly substantial on AIME 2025, where full reconstruction reaches 33.75\% versus 36.46\% for micro-reflective, and on HMMT 2025, where it reaches 18.75\% versus 20.62\%. These results validate the core hypothesis that preserving the learner's partial reasoning path is critical for learnability. Full reconstruction forces the model to learn an entirely new reasoning path rather than building upon its existing partial understanding. The gap is most pronounced on AIME 2024 at 4.38 points, followed by AIME 2025 at 2.71 points and HMMT 2025 at 1.87 points, suggesting that prefix preservation provides the largest benefit when the model has a moderate level of competence and the errors are more localized.

\subsection{Integration Component Ablation}
\label{sec:integration}

Incorporating constructed trajectories into RL training introduces distributional and optimization challenges. This section verifies the necessity of each integration component through ablation studies.

\begin{table*}[h]
\centering
\caption{Integration component ablation. Each row progressively adds components to the \method pipeline, from the most stripped-down variant (top) to the full \method (bottom).}
\label{tab:component_ablation}
\resizebox{\linewidth}{!}{
\begin{tabular}{l|cc|cc|cc}
\toprule
 & \multicolumn{2}{c|}{\textbf{AIME 2024}} & \multicolumn{2}{c|}{\textbf{AIME 2025}} & \multicolumn{2}{c}{\textbf{HMMT 2025}} \\
\textbf{Configuration} & \textbf{Pass@1} & \textbf{Pass@5} & \textbf{Pass@1} & \textbf{Pass@5} & \textbf{Pass@1} & \textbf{Pass@5} \\
\midrule
Cold-start & 31.67 & 51.04 & 23.75 & 37.08 & 14.37 & 26.46 \\
\quad + \method w/o DAE + w/o OTS + w/o Negative & 52.71 & 71.67 & 41.46 & 57.92 & 20.83 & 38.96 \\
\quad + \method w/o OTS + w/o Negative & 49.17 & 67.92 & 36.67 & 56.46 & \underline{21.67} & 42.92 \\
\quad + \method w/o OTS & \underline{57.29} & \underline{77.71} & \underline{45.83} & \underline{61.67} & 21.46 & \underline{46.25} \\
\quad + \textbf{\method} & \textbf{62.50} & \textbf{80.21} & \textbf{46.88} & \textbf{67.92} & \textbf{31.46} & \textbf{50.00} \\
\bottomrule
\end{tabular}
}
\end{table*}

Table~\ref{tab:component_ablation} progressively strips components from the \method pipeline. We discuss each component below, from the most impactful to the least, following the accumulation order in the table.

\paragraph{Negative Samples in the Reflection Group.} The reflection group in \method consists of original rollouts, constructed corrective trajectories, and negative samples. These negative samples are incorrect constructed trajectories that, despite undergoing the micro-reflective construction process, still fail to produce the correct answer. They serve as contrastive anchors during advantage estimation. Row~2 (w/o OTS + w/o Negative) removes negative samples from the reflection group while retaining decoupled advantage estimation. On AIME 2024 and AIME 2025, this causes a substantial performance drop relative to Row~1 (w/o DAE + w/o OTS + w/o Negative), with Pass@1 declining from 52.71 to 49.17 on AIME 2024 and from 41.46 to 36.67 on AIME 2025. On HMMT 2025, the direction is reversed, with Pass@1 improving slightly from 20.83 to 21.67, suggesting that the interaction between decoupling and negative samples is benchmark-dependent. The overall pattern reveals that decoupled advantage estimation without negative samples amplifies the bias toward corrective trajectories on benchmarks where the model has stronger baseline performance. When the reflection group contains only original trajectories and corrective ones, the group mean is artificially elevated by the consistently higher rewards of corrective trajectories, causing original correct samples to receive inappropriately low advantages. Negative samples serve as a counterweight, and their inclusion lowers the group mean, restoring the correct relative ordering of advantages and enabling \method to properly distinguish effective from ineffective corrections.

\paragraph{Decoupled Advantage Estimation.} \method computes advantages separately within original and reflection groups, rather than jointly over all samples. The effect of decoupled advantage estimation is isolated by comparing Row~1 (w/o DAE + w/o OTS + w/o Negative) and Row~2 (w/o OTS + w/o Negative). Row~2 adds decoupling while keeping negative samples and OTS removed. On AIME 2024 and AIME 2025, the performance drops from Row~1 to Row~2, a counterintuitive result that reveals the interaction between DAE and negative samples. On HMMT 2025, however, Row~2 slightly outperforms Row~1, indicating that the negative effect of decoupling without negative samples is less pronounced when the model's baseline accuracy is lower. When advantages are decoupled but negative samples are absent, the reflection group mean is artificially elevated by the consistently higher rewards of corrective trajectories. This causes advantages of original correct samples to be inappropriately low relative to the reflection group, creating a bias toward corrective trajectories at the expense of the original GRPO update. The gap from Row~1 to Row~3 captures the combined effect of DAE with negative samples, where the addition of negative samples in Row~3 restores the correct relative ordering of advantages and yields substantial recovery. This confirms the analysis in \S\ref{sec:decoupled_adv}. Without decoupling, corrective trajectories with high rewards inflate the joint group mean, causing original samples to receive disproportionately negative advantages. Since incorrect original samples tend to be longer than correct ones, this advantage contamination indirectly penalizes response length, driving entropy collapse.

\paragraph{OOD Token Suppression.} The OOD Token Suppression mechanism addresses token-level distributional mismatch, a finer-grained challenge than trajectory-level alignment. While self-construction maintains trajectory-level proximity (\S\ref{sec:reflective_ablation}), the reconstruction process may occasionally generate tokens that are distributionally distant from the learner's current policy. For instance, when the reconstruction explores reasoning patterns or vocabulary the model has not yet acquired, it creates a local distributional mismatch within an otherwise well-aligned trajectory. OTS weights each token's loss contribution by $w_t = \mathrm{clamp}(\exp(s_t), w_{\min}, w_{\max})$, where $s_t = \log p_\theta(y_t \mid \mathbf{x}, \mathbf{y}_{<t}) + H_t$ combines the token's log-probability under the current policy and the policy entropy. In-distribution tokens ($s_t \approx 0$) receive weight $\approx 1$, while OOD tokens ($s_t \ll 0$) are suppressed, preventing the model from being forced to learn far from its current distribution. OTS is applied only to constructed trajectories; original samples are unaffected.

The gap between Row~3 and \method isolates the contribution of OTS. Removing OTS degrades performance across all benchmarks, confirming that reconstructed trajectories contain OOD tokens that would harm learning under uniform weighting. The degradation is most pronounced on HMMT 2025, where longer chains provide more opportunities for the reconstruction process to introduce distributionally distant tokens. We analyze OTS dynamics in detail in \S\ref{sec:training_dynamics}.

\subsection{Training Dynamics}
\label{sec:training_dynamics}

\begin{figure}[h]
\centering
\includegraphics[width=\linewidth]{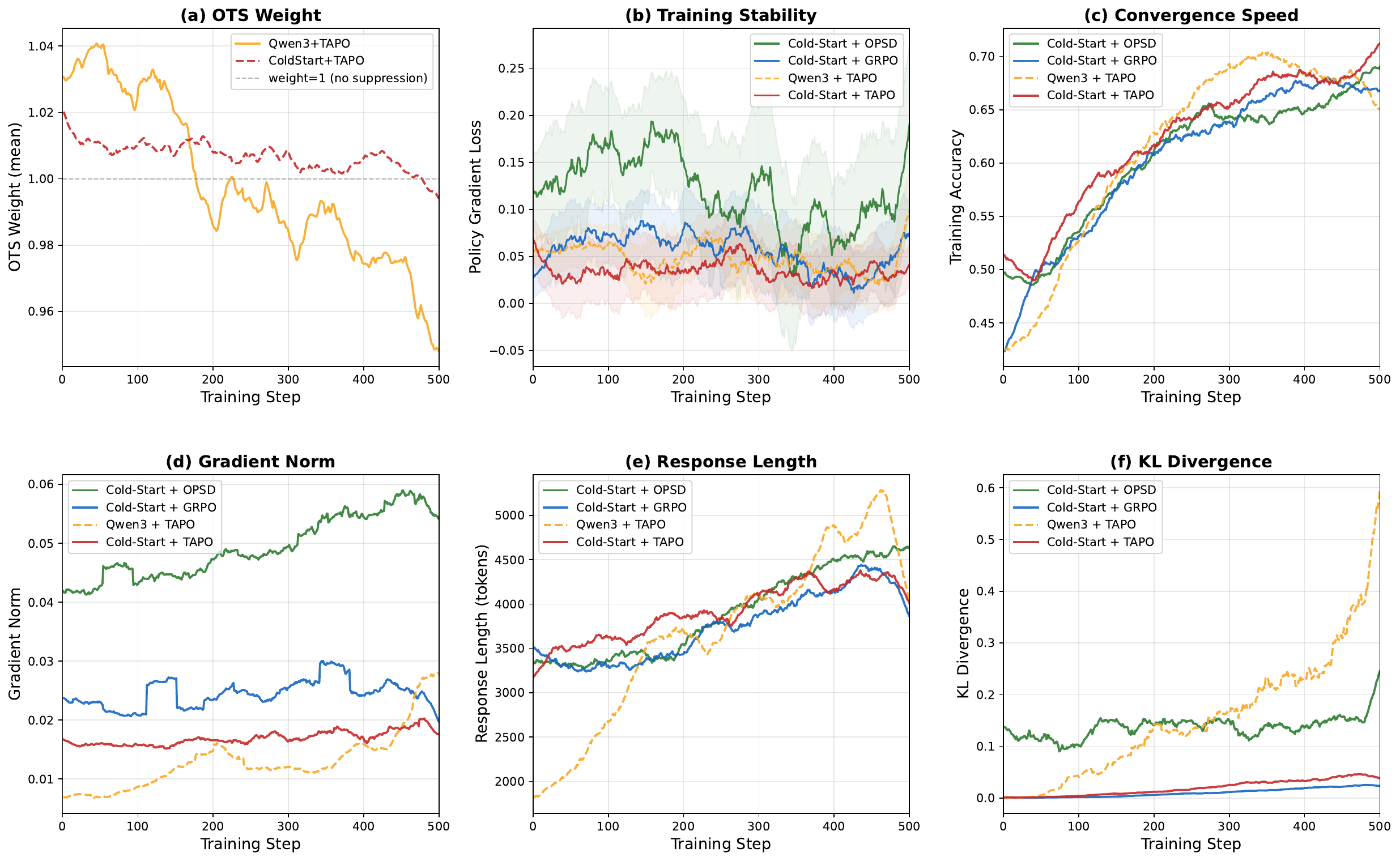}
\caption{Training dynamics comparing Cold-start+\method, Cold-start+GRPO, Cold-start+OPSD, and Qwen3+\method over 500 steps. (a) OTS weight for \method vs.~Qwen3+\method. (b) Policy gradient loss with rolling standard deviation. (c) Training accuracy. (d) Gradient norm. (e) Response length. (f) KL divergence between the current policy and the reference policy. All curves smoothed with a rolling window of 40$\sim$80 steps.}
\label{fig:training_dynamics}
\end{figure}

We analyze the training dynamics of \method over the course of one epoch. Figure~\ref{fig:training_dynamics} presents six complementary views of the training process, comparing \method against GRPO, OPSD, and the Qwen3+\method baseline without cold start.

\paragraph{OTS Weight.}
Figure~\ref{fig:training_dynamics}(a) tracks the evolution of OTS weight mean. Cold-start+\method maintains OTS weight close to 1.0 throughout training, indicating that the cold-start policy is well-aligned with the reconstruction distribution and requires minimal suppression. Qwen3+\method exhibits a gradual decline, with weight dropping to approximately 0.95 by step 500, reflecting that without cold start, the reconstruction process introduces tokens increasingly distant from the immature policy's distribution. This persistent gap demonstrates that cold start is essential for distributional alignment, and it pre-equips the policy with trajectory construction capability, reducing the OOD token burden that OTS must otherwise suppress.

\paragraph{Training Stability.}
Figure~\ref{fig:training_dynamics}(b) compares policy gradient loss. 
Cold-start and Qwen3+\method achieve lower pg\_loss than Cold-start+GRPO and Cold-start+OPSD, while the OPSD baseline exhibits the highest pg\_loss and the largest variance. The high volatility of OPSD reflects the distributional mismatch between the privileged distillation target and the learner's policy, which introduces noisy gradient signals. \method avoids this instability through decoupled advantage estimation, which computes advantages within distributionally coherent groups rather than mixing samples with different reward distributions.

\paragraph{Convergence Speed.}
Figure~\ref{fig:training_dynamics}(c) plots training accuracy. We note that training accuracy reflects the reward model's scoring and should be interpreted as a training signal indicator rather than a measure of final model capability. All configurations exhibit increasing training accuracy, with Cold-start+OPSD and Cold-start+\method tracking slightly higher than Cold-start+GRPO and Qwen3+\method in the mid-to-late training phase. The stronger training signal under \method is consistent with the final performance improvements reported in Table~\ref{tab:main_results}.

\paragraph{Gradient Norm.}
Figure~\ref{fig:training_dynamics}(d) tracks gradient norm. Cold-start+\method and Qwen3+\method maintain the lowest gradient norms, while Cold-start+OPSD exhibits the highest. Lower gradient norms indicate a smoother optimization landscape with fewer destabilizing updates. The well-conditioned optimization under \method is a direct consequence of decoupled advantage estimation and OTS weighting, which together prevent the gradient contamination that arises from mixing trajectories with heterogeneous reward distributions.

\paragraph{Response Length.}
Figure~\ref{fig:training_dynamics}(e) shows mean response length. All configurations exhibit increasing response length over training. Cold-start+OPSD shows the most pronounced growth, while Cold-start+\method and Cold-start+GRPO grow more moderately. The more controlled growth under \method and GRPO suggests that both methods avoid the runaway length inflation that can occur when the policy is pushed toward a privileged distribution without proper regularization.

\paragraph{KL Divergence.}
Figure~\ref{fig:training_dynamics}(f) tracks KL divergence. Cold-start+\method maintains low KL divergence, comparable to Cold-start+GRPO and substantially lower than Cold-start+OPSD. This indicates that training on constructed trajectories does not cause excessive policy drift when the cold-start initialization is used. In contrast, Qwen3+\method exhibits the highest KL divergence, nearly an order of magnitude above Cold-start+\method. The stark contrast between the two \method configurations underscores the necessity of cold start, i.e. without pre-alignment to the trajectory construction format, the policy is forced toward a distribution far from its initialization, resulting in aggressive KL growth. With cold start, the policy is already equipped to process constructed trajectories, and the combination of DAE and OTS keeps subsequent RL updates within a well-conditioned region.

\subsection{Difficulty-Aware Candidate Selection}
\label{sec:zpd_quantity}

The candidate selection mechanism (\S\ref{sec:candidate_selection}) is governed by the ZPD threshold $(n_{\text{pos}}, n_{\text{neg}})$ and the maximum construction count $m_{\text{max}}$. Table~\ref{tab:zpd_quantity} examines the sensitivity of \method to variations in these parameters. The default setting $(2, 4, 4)$ achieves the best Pass@1 on AIME 2024 and HMMT 2025, and the best Pass@5 on all three benchmarks. Relaxing the threshold to $(2, 2, 2)$ admits more queries into the ZPD by lowering the incorrect sample requirement from 4 to 2, but the reduced $m_{\text{max}}$ limits constructions per query to 2, causing a noticeable performance drop across all benchmarks. Increasing $m_{\text{max}}$ back to 4 while keeping the relaxed threshold, i.e., configuration $(2, 2, 4)$, recovers substantial performance, confirming that construction volume is an important factor. Tightening the threshold to $(1, 5, 3)$ restricts construction to a narrower set of high-confidence queries and yields the best Pass@1 on AIME 2025 while remaining competitive on HMMT 2025, though it underperforms on AIME 2024. The results show that \method is reasonably robust to moderate variations in these parameters. The default setting $(2, 4, 4)$ performs consistently well across benchmarks, though a stricter threshold may offer a slight advantage on specific benchmarks such as AIME 2025.

\begin{table*}[h]
\centering
\caption{Impact of ZPD threshold $(n_{\text{pos}}, n_{\text{neg}})$ and construction quantity $m_{\text{max}}$ across benchmarks. 
}
\label{tab:zpd_quantity}
\begin{tabular}{ccc|cc|cc|cc}
\toprule
 & & & \multicolumn{2}{c|}{\textbf{AIME 2024}} & \multicolumn{2}{c|}{\textbf{AIME 2025}} & \multicolumn{2}{c}{\textbf{HMMT 2025}} \\
$n_{\text{pos}}$ & $n_{\text{neg}}$ & $m_{\text{max}}$ & \textbf{Pass@1} & \textbf{Pass@5} & \textbf{Pass@1} & \textbf{Pass@5} & \textbf{Pass@1} & \textbf{Pass@5} \\
\midrule
2 & 2 & 2 & 51.88 & 73.75 & 37.92 & 56.25 & 22.08 & 39.79 \\
2 & 2 & 4 & \underline{57.92} & \underline{78.12} & 40.21 & 58.75 & 24.17 & 42.92 \\
2 & 4 & 4 & \textbf{62.50} & \textbf{80.21} & \underline{46.88} & \textbf{67.92} & \textbf{31.46} & \textbf{50.00} \\
1 & 5 & 3 & 57.08 & 76.46 & \textbf{48.96} & \underline{66.04} & \underline{30.42} & \underline{48.96} \\
\bottomrule
\end{tabular}
\end{table*}

%% file: sections/60-conclusion.tex
\section{Conclusion}
Current self-distillation methods improve reasoning through implicit distributional alignment, treating errors as something to be suppressed rather than leveraged. In this paper, We proposed \textbf{T}rajectory-\textbf{A}ugmented \textbf{P}olicy \textbf{O}ptimization (\method), which advances self-distillation from implicit alignment to explicit trajectory construction. \method constructs micro-reflective corrections from the model's own error patterns, integrated through difficulty-aware ZPD candidate selection, decoupled advantage estimation, and OOD token suppression. Experiments on AIME 2024, AIME 2025, and HMMT 2025 demonstrate consistent improvements over GRPO and OPSD, with DSR and ERR analysis confirming that \method strengthens both first-pass reasoning and error-correction effectiveness. Ablation studies validate the contribution of each component.

%% file: sections/70-acknowledge.tex
\section*{Acknowledgments}
This work was supported by Qwen Business Unit through Alibaba Research Intern Program